\documentclass[
]{ceurart}

\usepackage{listings}
\usepackage{inconsolata}
\usepackage{booktabs,arydshln}
\usepackage{graphicx}
\usepackage{amsmath}
\usepackage{hyperref}
\usepackage{amssymb}
\usepackage{arydshln}
\usepackage{tikz}
\usepackage{xcolor}
\usepackage{pgfplots}
\usepackage{caption}
\usepackage{tabularx,array}
\usepackage{fnpct}
\usepackage{booktabs,xltabular}
\usepackage{adjustbox}

\makeatletter
\def\adl@drawiv#1#2#3{%
        \hskip.5\tabcolsep
        \xleaders#3{#2.5\@tempdimb #1{1}#2.5\@tempdimb}%
                #2\z@ plus1fil minus1fil\relax
        \hskip.5\tabcolsep}
\newcommand{\cdashlinelr}[1]{%
  \noalign{\vskip\aboverulesep
           \global\let\@dashdrawstore\adl@draw
           \global\let\adl@draw\adl@drawiv}
  \cdashline{#1}
  \noalign{\global\let\adl@draw\@dashdrawstore
           \vskip\belowrulesep}}
\makeatother

\definecolor{color1}{RGB}{159,168,218}
\definecolor{color2}{RGB}{173,201,198}
\definecolor{color3}{RGB}{207,216,220}

\definecolor{color4}{RGB}{247,194,183}
\definecolor{color5}{RGB}{187,230,190}
\definecolor{color6}{RGB}{187,230,222}
\definecolor{darkgreen}{rgb}{0,0.5,0}

\begin{document}

\copyrightyear{2023}
\copyrightclause{Copyright for this paper by its authors.
  Use permitted under Creative Commons License Attribution 4.0
  International (CC BY 4.0).}


\conference{ISWC 2023: Deep Learning for Knowledge Graph(DL4KG), November 06-10, 2023, Athens, Greece}

\title{NNKGC: Improving Knowledge Graph Completion with Node Neighborhoods}




\author[1]{Irene Li}[%
orcid=0000-0002-1851-5390,
email=ireneli@ds.itc.u-tokyo.ac.jp,
]
\cormark[1]
\address[1]{The University of Tokyo, Tokyo, Japan}

\author[1]{Boming Yang}[%
orcid=0009-0004-6298-5711,
email=boming@g.ecc.u-tokyo.ac.jp
]

\cortext[1]{Corresponding author.}

\begin{abstract}
Knowledge graph completion (KGC) aims to discover missing relations of query entities. Current text-based models utilize the entity name and description to infer the tail entity given the head entity and a particular relation. Existing approaches also consider the neighborhood of the head entity. However, these methods tend to model the neighborhood using a flat structure and are only restricted to 1-hop neighbors. In this work, we propose a node neighborhood-enhanced framework for knowledge graph completion. It models the head entity neighborhood from multiple hops using graph neural networks to enrich the head node information. Moreover, we introduce an additional edge link prediction task to improve KGC. Evaluation on two public datasets shows that this framework is simple yet effective. The case study also shows that the model is able to predict explainable predictions.
\end{abstract}


\maketitle

\section{Introduction}
A knowledge graph (KG) is a structured representation of entities, their properties, and their relations \cite{ehrlinger2016towards}. It has become increasingly popular in recent years due to its ability to support various applications, including question answering, recommendation systems, and semantic search \cite{steiner2012adding, Wang2021LearningIB,Liang2022KG4PyAT}. However, a KG is usually defined by domain experts, so a few relations might be missing. Knowledge Graph Completion (KGC) aims to discover these missing relations in a given KG \cite{Lin2015LearningEA}.

The methods for KGC mainly fall into two groups: embedding-based and text-based. Embedding-based methods consider the knowledge graph relations as graph structures, then learn low-rank embeddings to represent entities and relations, such as TransE \cite{Bordes2013TranslatingEF}, TuckER  \cite{balazevic-etal-2019-tucker} and so on. However, such methods fail in any inductive scenarios. Text-based methods infer relations based on entities and the corresponding descriptions through representation learning to solve this problem \cite{Yao2019KGBERTBF,kim-etal-2020-multi,Safavi2021RelationalWK}. KEPLER \cite{wang-etal-2021-kepler} utilizes pretrained language modeling representation with knowledge embedding to integrate factual knowledge for KGC. The recent SimKGC \cite{Wang2022SimKGCSC} model applies a contrastive learning approach with new methods for negative sampling, outperforming several embedding-based methods.

Most existing work typically models the head entity $h$ and the query relation $r$ together, then focuses on the modeling for predicting the relation for the tuple $(h,r,?)$ to find the correct tail entity $t$.  The modeling for $(h,r)$ tends only to consider the head entity name or descriptions and the query relation name.  Besides, existing approaches also consider the neighborhood of the head entity \cite{Zhuo2022ANF,Borrego2021CAFEKG}. 
However, these methods tend to model the neighborhood using a flat structure so that a simple attention-based method can be applied; or they are only restricted to 1-hop neighbors for computational consideration. 
In this work, we propose considering the neighborhood from \textit{multiple hops} to improve the richness of the head entity information for further KGC predictions. By doing so, the head entity contains much explicit information that may help the relation prediction, especially when the head entity name/description is short and not informative. We model the neighborhood using a graph neural network, making it easy to propagate information within multiple hops away. 
Besides, to further enhance the correlation with these neighbors and the tail entity, we add an extra link prediction loss to the KGC task, and we show this could improve the KGC performance.

Our main contribution of this work is to propose a node neighborhood-enhanced model for knowledge graph completion by modeling the neighbors with graph neural networks \cite{Wu2019ACS}. Moreover, the experiment shows that this framework is effective in two public datasets. Additional case studies show that this model could provide explainable predictions. We release our code at \url{https://github.com/IreneZihuiLi/NNKGC}.

\section{Method}
\begin{figure}[t]
\centering
\includegraphics[width=12cm]{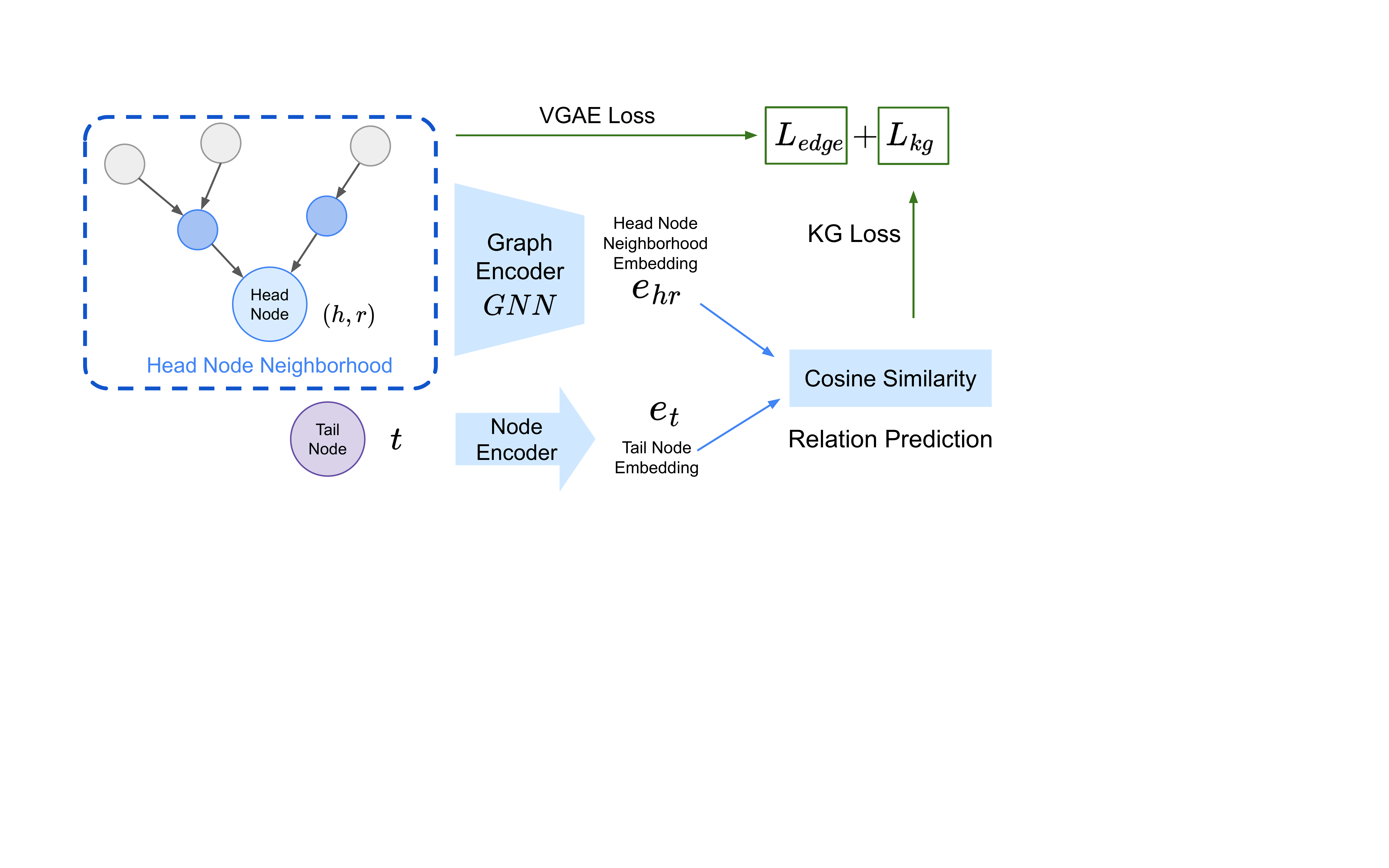}
\caption{NNKGC Model Illustration: we initially identify the neighborhood of the head node. Following this, we employ a graph encoder to encode both the head node and its corresponding relation. Simultaneously, the tail node is encoded. For the final link prediction, we utilize cosine similarity. A novel aspect of our methodology is the introduction of the VGAE loss. This is optimized in conjunction with the knowledge graph task loss during the model's training phase.}
\label{fig:kgc_model}
\end{figure}

We propose the \textbf{N}ode \textbf{N}eighborhood-enhanced model for \textbf{k}nowledge \textbf{G}raph \textbf{C}ompletion (NNKGC), which consists of the node encoder and the relation prediction module, as shown in Fig.~\ref{fig:kgc_model}. Following the setting of the previous work \cite{Wang2022SimKGCSC,Yao2019KGBERTBF}, we define the task to be the following: given an incomplete knowledge graph $\mathcal{G}$, we need to find the tail entity $t$ by providing the head entity $h$ and the query relation $r$: $(h,r,?)$.

\textbf{Node Encoder} Typically, we model the head and tail entity nodes separately. In our case, each entity node contains a node name in a short phrase or a keyword (\textit{name}), and a node description in a sentence  (\textit{desc}). The relation $r$ is usually represented by a phrase or a keyword (\textit{rel}). To incorporate the text features into the knowledge graph, we utilize the pre-trained language models (PLMs), BERT \cite{devlin-etal-2019-bert}, to encode the head entity and relations. 
Each entity node is represented by: 
\begin{equation}
e_{node} = \mathbf{BERT}[name,desc, \texttt{[SEP]}, rel]
\end{equation}
To further consider the head node neighborhood to enrich the encoding, we apply a graph neural network \cite{scarselli2009graph} to model the neighbors. We first collect all the neighbors within $k$-hop as a list of nodes: $nb1, nb2,...nbl$. Then we encode the head entity and relation as follows: 
\begin{align}
X &= [e_{head}, e_{nb1},e_{nb2},...e_{nbl}],  \\
e_{hr} &= \mathbf{GNN}(X,A),
\label{eq:graph} 
\end{align}

In which, $X$ denotes the node embeddings of the entities within the neighborhood, including the head node, as well as the neighbors, and $A$ is the adjacency matrix indicating the directed connections among the entities. We define the neighborhood of a head node to be the connected entities, including incoming and outgoing connections. In Fig ~\ref{fig:kgc_model}, we illustrate 1-hop (blue) and 2-hop (grey) neighbors and ignore outgoing neighbors for simplicity. There is a number of choices for the graph encoder, we investigate three types: graph convolutional network (GCN) \cite{Kipf2016SemiSupervisedCW}, graph attention network (GAT) \cite{Velickovic2017GraphAN}, and GraphSAGE \cite{Hamilton2017InductiveRL}. 

\textbf{Relation Prediction} To make the prediction on the tail entities, we followed the recent SimKGC \cite{Wang2022SimKGCSC} model. SimKGC introduces a contrastive learning approach to improve the negative sampling and applies the InfoNCE loss as the loss function. The prediction is simply a score based on cosine similarity between $e_{hr}$ and $e_t$. Then the tail entities are ranked by this score and the largest one is chosen as the prediction. The loss for the tail entity prediction is denoted as ${L}_{kg}$.

\textbf{Neighborhood Edge Prediction} To further strengthen the effect of neighborhood entities for the KGC task, we ask the model to predict neighborhood edges (i.e., the highlighted edge in the figure). In other words, we randomly mask some edges in matrix $A$, and let the model reconstruct these masked relations. There are multiple ways to achieve this, and we apply a variational graph autoencoder (VGAE) \cite{li2021unsupervised} to conduct the missing link prediction. It is shown to be effective for several NLP tasks \cite{li2021unsupervised,Xie2021InductiveTV}. Specifically, we apply vanilla GCNs to model the mean and standard deviation of the node features, then conduct dot-product to predict the existence of a given node pair. Note that this is only done for the neighborhood. It will generate a loss on the reconstructed edges $\mathcal{L}_{edge}$. So the final loss becomes:
\begin{equation}
\mathcal{L} = \lambda\mathcal{L}_{kg} +  (1-\lambda) \mathcal{L}_{edge} 
\end{equation}

Here, $\lambda$ is an empirical constant to control the effect of the edge prediction, and it ranges between 0 and 1.

\begin{table*}[htbp]
\caption{Dataset statistics: \texttt{vocab} indicates the vocabulary size; \texttt{node mean} indicates the average number of tokens for the node descriptions, and the \texttt{node median} indicates the median number of tokens.}
\label{tab:ds}
\centering
\begin{adjustbox}{width=\textwidth}
\begin{tabular}{ccccccccc} \toprule
\textbf{dataset} & \textbf{entity} & \textbf{relation} & \textbf{mean} & \textbf{median} & \textbf{train/valid/test} & \textbf{vocab} & \textbf{node mean} & \textbf{node median} \\ \midrule
WN18RR    & 40,943 & 11       & 3.7  & 2      & 86, 835/3,034/3,134   & 18,741  & 17.1      & 15          \\
FB15k-237 & 14,541 & 237      & 2.8  & 2      & 272,115/17,535/20,466 & 28,538 & 114.6     & 76         \\ \bottomrule
\end{tabular}
\end{adjustbox}
\end{table*}

\section{Experiment}

Shown in Tab.~\ref{tab:ds}, we conduct knowledge graph completion on two public datasets: FB15k-237 \cite{Toutanova2015RepresentingTF}, and WN18RR \cite{Dettmers2017Convolutional2K}. We follow the previous work \cite{wang-etal-2021-kepler,Wang2022SimKGCSC} to report the following evaluation metrics: mean reciprocal rank (MRR), Hits@{1,3,10}. 

\textbf{Hyperparameters}
The BERT encoder is the pretrained \textit{bert-base-uncased} model \footnote{\url{https://huggingface.co/bert-base-uncased}}. AdamW optimizer \cite{Loshchilov2017DecoupledWD} was applied for training. For modeling the neighborhood, we typically apply two layers of neural networks, and the dimension for the graph node representation is 768. We set the number of attention heads for the GAT encoder to 3. We set the $\lambda$ to 0.2. 
We conduct experiments on 4 A100 GPUs (with 40GB memory). Training on WN18RR with 30 epochs took about 3 hours; training on FB15k237 with 5 epochs took less than one hour.
More experimental hyperparameters can be found in the code URL.

\begin{table*}[htbp]
\caption{Main results on selected datasets. We report MRR, H@1, H@3 and H@10 scores. }
\centering
\begin{tabular}{crrrr} \toprule
\textbf{FB15k-237} & \textbf{MRR} & \textbf{H@1} & \textbf{H@3} & \textbf{H@10} \\\midrule
TransE \cite{Bordes2013TranslatingEF}             & 27.9         & 19.8         & 37.6         & 44.1                 \\
TuckER \cite{balazevic-etal-2019-tucker}             & \underline{35.8}         & \underline{26.6}         & \underline{39.4}         & \underline{54.4}                \\
MTL-KGC \cite{Kim2020MultiTaskLF}            & 26.7         & 17.2         & 29.8         & 45.8                \\
SimKGC \cite{Wang2022SimKGCSC}            & 33.6         & 24.9         & 36.2         & 51.1                 \\ 
\cdashlinelr{1-5}
NNKGC              & 33.2         & 24.6         & 35.9         & 50.4                 \\
NNKGC$_e$         & \textbf{33.8}	   &\textbf{25.2}	  &\textbf{36.5}	  &\textbf{ 51.5}	\\ \midrule \midrule
\textbf{WN18RR}    & \textbf{MRR} & \textbf{H@1} & \textbf{H@3} & \textbf{H@10} \\ \midrule
TransE \cite{Bordes2013TranslatingEF}             & 24.3         & 4.3          & 44.1         & 53.2              \\
TuckER \cite{balazevic-etal-2019-tucker}             & 47.0         & 44.3         & 48.2         & 52.6               \\
MTL-KGC \cite{Kim2020MultiTaskLF}           & 33.1         & 20.3         & 38.3         & 59.7             \\
SimKGC \cite{Wang2022SimKGCSC}             & \underline{66.6}         & \underline{58.7}         & \underline{71.7}         & \underline{80.0}                \\  \cdashlinelr{1-5}
%
NNKGC              & 67.3         & 59.8         & 71.7         & 80.5               \\
NNKGC$_e$          &\textbf{67.4}       & \textbf{59.6}        & \textbf{72.2}         & \textbf{81.2}         \\ \bottomrule
\end{tabular}
\label{tab:res}
\end{table*}

\textbf{Main Results}
We compare with two embedding-based methods: TransE \cite{Bordes2013TranslatingEF}, and TuckER \cite{balazevic-etal-2019-tucker}. TransE learns low-dimensional embeddings of the entities, and TuckER applies Tucker decomposition of the binary tensor representation for knowledge graph triples. We also include text-based methods for baselines. MTL-KGC \cite{Kim2020MultiTaskLF} is a multi-task learning method for KGC, and SimKGC \cite{Wang2022SimKGCSC}. We compare two settings of our NNKGC model: without the edge loss $\mathcal{L}_{edge}$ (NNKGC) and with the edge loss (NNKGC$_{e}$). Text-based methods performs better than embedding-based methods. Without the VGAE loss, our NNKGC is competitive with the best baseline, SimKGC. 
In general, our best model NNKGC$_{e}$ outperforms the selected baselines in both datasets.

\section{Ablation Study}

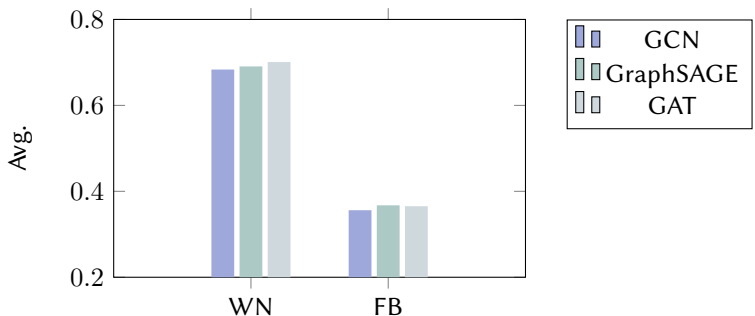
\begin{figure}[htbp]
\centering
\begin{tikzpicture}
\begin{axis}[  ybar,  ymin=0.2,  ymax=0.8,  bar width=0.3cm,   ylabel={Avg.},  symbolic x coords={WN, FB},  xtick=data,  enlarge x limits=1,  
legend style={at={(1.1,1)},anchor=north west,font=\small},
height=5cm,width=7cm,]
\addplot[fill=color1,draw=none] coordinates {(WN,0.6834) (FB,0.355975)};
\addplot[fill=color2,draw=none] coordinates {(WN,0.690725) (FB,0.3674)};
\addplot[fill=color3,draw=none] coordinates {(WN,0.70085) (FB,0.365275)};
\legend{ GCN, GraphSAGE, GAT}
\end{axis}
\end{tikzpicture}
\caption{Comparison of multiple neighborhood graph modeling: GCN, GraphSAGE and GAT. }
\label{fig:abl}
\end{figure}

\textbf{Neighborhood Encoder}
We then study different neighborhood graph encoders in Eq.~\ref{eq:graph}. We compare three widely-used graph models: GCN, GAT, and GraphSAGE. In Fig.~\ref{fig:abl}, we report the average performance on the two datasets.  We could observe that GraphSAGE and GAT are slightly better than the vanilla GCN encoder. 

We provide a detailed evaluation of various graph encoders for modeling the head node neighborhood, shown in Tab.~\ref{app_res_wn} and ~\ref{app_res_fb}. In general, the GCN encoder is worse than the other two. In general, GraphSAGE shows better stability in both datasets. 

\begin{table}[h]
\caption{Comparison of multiple neighborhood graph modeling: GCN, GraphSAGE, and GAT on the FB15k-237 dataset.}
\centering
\begin{tabular}{crrrr} \toprule
\textbf{Encoder} & \textbf{MRR} & \textbf{H@1} & \textbf{H@3} & \textbf{H@10} \\ \midrule
GCN         & 32.7         & 23.8         & 35.4         & 50.6          \\
GraphSAGE        & \textbf{33.8}         & \textbf{25.2}         & \textbf{36.5}         & \textbf{51.5}          \\
GAT         & 33.7         & 25.1         & 36.2         & 51.2          \\ \bottomrule
\end{tabular}
\label{app_res_fb}
\end{table}

\begin{table}[h]
\caption{Comparison of multiple neighborhood graph modeling: GCN, GraphSAGE, and GAT on the WN18RR dataset.}
\centering
\begin{tabular}{crrrr} \toprule
\textbf{Encoder}        & \textbf{MRR}  & \textbf{H@1}  & \textbf{H@3}  & \textbf{H@10 } \\ \midrule
GCN       & 65.4 & 57.0 & 70.7 & 80.2 \\
GraphSAGE & 65.8 & 56.4 & \textbf{72.4} & 81.7 \\
GAT       & \textbf{67.4} & \textbf{59.6 }& 72.2 & \textbf{81.2} \\ \bottomrule
\end{tabular}
\label{app_res_wn}
\end{table}

\begin{table*}[t]
\caption{An example from FB15k-237: we show the ground truth tail and predictions, as well as a selection of neighbor entities. We highlight \colorbox{color5}{direct} triggering tokens and \colorbox{color6}{other relevant} tokens. }
	\centering
	\small
	\begin{tabularx}{\textwidth}{X}
	\toprule 
\textbf{Head:} Star Wars Episode IV: A New Hope \\
\textbf{Head node description:} \textit{ Star Wars, later retitled Star Wars Episode IV: A New Hope, is a 1977 American epic space opera film written and directed by George Lucas...} \\ 
\textbf{Relation: }nominated for (an award) \\
\textbf{Ground Truth Tail:} Academy Award for Best Sound Mixing \\
\textbf{Predicted Tails:} \textcolor{darkgreen}{Academy Award for Best Sound Mixing (0.71)}, BAFTA Award for Best Special Visual Effects (0.656), Golden Globe Award for Best Original Score (0.65) \\ \midrule
\textbf{Neighbor 1, Name:} \colorbox{color5}{Academy Award} for Best Production Design\\ 
\textbf{Description}: \textit{\colorbox{color5}{The Academy Awards} are the oldest awards ceremony for achievements in motion pictures. \colorbox{color6}{The Academy Award for Best Production Design} recognizes achievement in art direction on a film. The category's original name was Best Art Direction, but was changed to its current name in 2012 for the 85th Academy Awards.} \\
\textbf{Node relation:} \colorbox{color5}{nominated for (an award)}
 \\ \cdashlinelr{1-1}
\textbf{Neighbor 2, Name:} Ben Burtt \\
\textbf{Description}: \textit{Benjamin Ben Burtt, Jr. is an American \colorbox{color5}{sound designer}, film editor, director, screenwriter, and voice actor. He has worked as sound designer on various films including: the \colorbox{color5}{Star Wars} and Indiana Jones film series...} \\
\textbf{Node relation:} \colorbox{color5}{winner, won (an award) }
  \\ \cdashlinelr{1-1}
\textbf{Neighbor 3, Name:} John Williams \\
\textbf{Description}: \textit{John Towner Williams is an American \colorbox{color6}{composer}, conductor and pianist...}\\
\textbf{Node relation:} \colorbox{color6}{music film}
  \\\bottomrule
	\end{tabularx}
\label{tab:case1}
\end{table*}

\textbf{Neighborhood Hops}
As we focus on the neighborhood, we now study the effect of the neighborhood scale. We conduct experiments on 1-hop, 2-hop and 3-top neighbors to the head entity. Shown in Fig.~\ref{fig:hop_wn}, we compare the evaluation metrics as well as the average performance (\texttt{Avg.} on the WN18RR dataset. 

\begin{figure}[htbp]
\caption{The effect of hop numbers, results on WN18RR. }
    \centering
    \begin{tikzpicture}
\begin{axis}[    xlabel={Hop Number},  ymin=0.5, ymax=0.9,    xmin=0, xmax=4,    xtick={1,2,3},    ytick={0,0.2,0.4,0.6,0.8,1.0},    grid=major,    grid style={dashed,gray!30},    at={(0,0)},    anchor=south west,
legend style={at={(1.1,1)},anchor=north west,font=\small},
height=4cm,width=6cm,
]

\addplot[color=blue,mark=o,]
    coordinates {(1,0.6737)(2,0.6511)(3,0.6409)};
\addlegendentry{MRR}

\addplot[color=red,mark=square,]
    coordinates {(1,0.5960)(2,0.5563)(3,0.5407)};
\addlegendentry{H@1}

\addplot[color=color1,mark=triangle,]
    coordinates {(1,0.7220)(2,0.7135)(3,0.7082)};
\addlegendentry{H@3}

\addplot[color=purple,mark=diamond,]
    coordinates {(1,0.8117)(2,0.8153)(3,0.8154)};
\addlegendentry{H@10}

\addplot[color=brown,mark=star, style=very thick,]
    coordinates {(1,0.7009)(2,0.6841)(3,0.6763)};
\addlegendentry{Avg.}
\end{axis}
\vspace{-4mm}
\end{tikzpicture}
    
    \label{fig:hop_wn}
\end{figure}

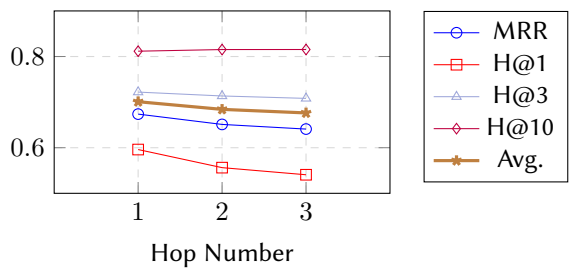
\begin{figure}[htbp]
\caption{Effect of the Number of Hops, results on FB15k-237.}
\centering
\begin{tikzpicture}
\begin{axis}[    xlabel={Number of Hop},     ymin=0.2, ymax=0.4,    xmin=0, xmax=4,    xtick={1,2,3},    ytick={0.2,0.3,0.4},    grid=major,    grid style={dashed,gray!30},    at={(0,0)},    anchor=south west,
legend style={at={(1.1,1)},anchor=north west,font=\small},
height=4cm,width=6cm,
]

\addplot[color=blue,mark=o,]
    coordinates {(1,0.3382)(2,0.3330)(3,0.3254)};
\addlegendentry{MRR}

\addplot[color=red,mark=square,]
    coordinates {(1,0.2517)(2,0.2436)(3,0.2363)};
\addlegendentry{H@1}

\addplot[color=color1,mark=triangle,]
    coordinates {(1,0.3652)(2,0.3608)(3,0.3551)};
\addlegendentry{H@3}

\addplot[color=purple,mark=diamond,]
    coordinates {(1,0.5146)(2,0.5121)(3,0.5046)};
\addlegendentry{H@10}

\addplot[color=brown,mark=star,]
    coordinates {(1,0.3674)(2,0.3624)(3,0.3554)};
\addlegendentry{AVG}
\end{axis}
\end{tikzpicture}
\label{fig:app_hop_fb}
\end{figure}

 We also show the bar chart of how the number of hops affects the performance for FB15k-237 in Fig.~\ref{fig:app_hop_fb}. One may observe a similar trend to that of WN18RR. As more hops will bring more neighboring entities, some noises may be added, which may make the performance worse.

In general, integrating more neighbors is not contributing to the performance. Especially, H@1 drops significantly when the hop number increases. A possible reason might be that more neighbors may introduce more noises which confuses the knowledge graph completion task. Moreover, doing so brings more computational burden to the model as more nodes participate in the graph modeling part. We did not conduct experiments on 4 and more hops, as the total training time is incredibly long.

\begin{table*}[t]
\caption{An example from WN18RR: we show the ground truth tail and predictions, as well as a selection of neighbor entities. We highlight \colorbox{color5}{direct} triggering tokens and \colorbox{color6}{other relevant} tokens. }
	\centering
	\small
	\begin{tabularx}{\textwidth}{X}
	\toprule 
\textbf{Head:} grant (NN) \\
\textbf{Head node description:} \textit{any monetary aid} \\ 
\textbf{Relation: }hypernym \\
\textbf{Ground Truth Tail:} financial aid (NN1) \\
\textbf{Predicted Tails:} \textcolor{darkgreen}{financial aid (NN1) (0.81)}, grant (NN1) (0.68), foreign aid (NN1)(0.572),
\\ \midrule
\textbf{Neighbor 1, Name:} \colorbox{color5}{grant} in aid (NN2) \\ 
\textbf{Description}: \textit{a grant to a person or school for some educational project} \\
\textbf{Node relation:} \colorbox{color5}{hypernym}
 \\ \cdashlinelr{1-1}
\textbf{Neighbor 2, Name:} grant (VB2) \\
\textbf{Description}: \textit{give as judged due or on the basis of merit; the referee awarded a free kick to the team; the jury awarded a million dollars to the plaintiff; \colorbox{color6}{funds are granted to qualified researchers}.} \\
\textbf{Node relation:} derivationally related form
  \\ \bottomrule
	\end{tabularx}
\label{tab:case2}
\end{table*}

\begin{table*}[t]
\caption{An example from FB15k-237: we show the ground truth tail and predictions, as well as a selection of neighbor entities. We highlight \colorbox{color5}{direct} triggering tokens and \colorbox{color6}{other relevant} tokens. }
	\centering
	\small
	\begin{tabularx}{\textwidth}{X}
	\toprule 
\textbf{Head:} Jack London \\
\textbf{Head node description:} \textit{John Griffith Jack London was an \colorbox{color5}{American} author, journalist, and social activist. He was a pioneer in the then-burgeoning world of commercial magazine fiction and was one of the first fiction writers to obtain worldwide celebrity and a large fortune from his fiction alone...} \\ 
\textbf{Relation: }nationality person people \\
\textbf{Ground Truth Tail:} United States of America \\
\textbf{Predicted Tails:} \textcolor{darkgreen}{United States of America (0.91)}, Confederate States of America (0.40), Union (0.34),
\\ \midrule
\textbf{Neighbor 1, Name:} San Francisco \\ 
\textbf{Description}: \textit{San Francisco, officially the City and County of San Francisco, is the leading financial and cultural center of Northern California and the San Francisco Bay Area...} \\
\textbf{Node relation:} place of birth person people
 \\ \cdashlinelr{1-1}
\textbf{Neighbor 2, Name:} \colorbox{color6}{Oakland} \\
\textbf{Description}: \textit{Oakland, located in the \colorbox{color5}{U.S. state of California}, is a major West Coast port city and the busiest port for San Francisco Bay and all of Northern California. It is the third largest city in the San Francisco Bay Area, the eighth-largest city in the state, and the 47th-largest city} \\
\textbf{Node relation:} \colorbox{color6}{location place, lived places}
  \\ \cdashlinelr{1-1}
\textbf{Neighbor 3, Name:} \colorbox{color6}{University of California, Berkeley} \\
\textbf{Description}: \textit{The University of California, Berkeley, is a public research university located in Berkeley, California, \colorbox{color5}{United States}. The university occupies 1,232 acres on the eastern side of the San Francisco Bay with the central campus resting on 178 acres. Berkeley is the flagship institution of the 10 campus University of California} \\
\textbf{Node relation:}\colorbox{color6}{students, graduates educational institution}
  \\ \bottomrule
\end{tabularx}
\label{tab:case3}
\end{table*}

\textbf{Case Study}
We select an example from FB15k-237 in Tab.~\ref{tab:case1}, and study how the neighbor entities and corresponding relations help with the KGC  task. Given the head entity to be the movie \textit{Star Wars IV}, and the relation to be \textit{nominated for (an award)}, our model predicts a ranked possible tail entity list. The entity \textit{Academy Award for Best Sound Mixing} achieves the highest score, which is precisely the ground truth. We also randomly select three 1-hop neighbors, and their relations with the head entity in the training set. As we can see, the first neighbor is one of the Academy Award categories, and the relation is the same as the query relation. The second neighbor is a sound designer. The description shows that this person worked on the \textit{Star Wars} series. The third neighbor is a composer who participates in the music production for the movie. We highlight some potential triggering tokens in the table. As we can see, some tokens are highly related to the ground truth tail entity, i.e., neighbor 1, the name \textit{ Academy Award}. Besides, the token \textit{composer} and \textit{music} are also semantically related to the ground truth tail entity name \textit{Best Sound Mixing}.  From this case study, one can notice that the neighborhood has some positive effects on the KGC task, which is consistent with our motivation. Moreover, we show that our framework has the potential to explain the predictions, which is an essential step for further fact verification.

We present two more case studies. Tab.~\ref{tab:case2} shows a random example from WN18RR. As we can see, the entity description tends to be shorter, but we can still find text chunks that are highly related to the ground truth tail entity (\textit{financial aid} and \textit{funds}). Similarly, Tab.~\ref{tab:case3} gives an example of finding a person's nationality. We can observe that the neighbors are some relevant locations about this head entity person (i.e., lives in \textit{Oakland}, \textit{U.S. state of California}), which contributes to the final prediction on the tail entity (\textit{United States of America}).





\section{Conclusion}

In this work, we proposed a node neighborhood-enhanced model for knowledge graph completion, by modeling the neighbors with graph neural networks. We showed that the framework is simple but effective. Case studies also show that it is possible to provide explainable results.


    







\bibliography{anthology,sample-ceur}

\end{document}